\pdfoutput=1

\documentclass[11pt]{article}

\usepackage[final]{acl}

\usepackage{times}
\usepackage{latexsym}

\usepackage[T1]{fontenc}

\usepackage[utf8]{inputenc}

\usepackage{microtype}

\usepackage{inconsolata}

\usepackage{graphicx}

\usepackage{booktabs}

\title{Merge to Learn: Efficiently Adding Skills to Language Models\\with Model Merging}

\author{
 \textbf{Jacob Morrison}\textnormal{\textsuperscript{1}}
 \enskip
 \textbf{Noah A. Smith}\textnormal{\textsuperscript{1,2}}
 \enskip
 \textbf{Hannaneh Hajishirzi}\textnormal{\textsuperscript{1,2}} 
 \AND
 \textbf{Pang Wei Koh}\textnormal{\textsuperscript{1,2}} 
 \enskip
 \textbf{Jesse Dodge}\textnormal{\textsuperscript{1}}
 \enskip
 \textbf{Pradeep Dasigi}\textnormal{\textsuperscript{1}}
 \AND
 \textnormal{\textsuperscript{1}Allen Institute for AI} \quad
 \textnormal{\textsuperscript{2}University of Washington} \AND
 \small{
   \textbf{Correspondence:} \href{mailto:jacobm@allenai.org}{jacobm@allenai.org}
 }
}

\begin{document}
\maketitle
\begin{abstract}
Adapting general-purpose language models to new skills is currently an expensive process that must be repeated as new instruction datasets targeting new skills are created, or can cause the models to forget older skills. In this work, we investigate the effectiveness of adding new skills to preexisting models by training on the new skills in isolation and later merging with the general model (e.g. using task vectors). In experiments focusing on scientific literature understanding, safety, and coding, we find that the \textit{parallel-train-then-merge} procedure, which is significantly cheaper than retraining the models on updated data mixtures, is often comparably effective. Our experiments also show that parallel training is especially well-suited for enabling safety features in LMs relative to continued finetuning and retraining, as it dramatically improves model compliance with safe prompts while preserving its ability to refuse dangerous or harmful prompts.
\end{abstract}

\section{Introduction}

Recent work has shown that instruction tuning pretrained language models (LMs) can result in strong generalist models that can perform a variety of comprehension and generation tasks, owing largely to the availability of high quality datasets.

As training datasets targeting new skills are constructed, it is an open question how best to \textit{patch} preexisting models to incorporate the new skills represented by those datasets. Commonly used approaches include continued finetuning (CFT) of the existing models on the new datasets and retraining (RT) the models on a combination of old and new instruction tuning datasets, both of which have clear pros and cons associated with them. 

CFT, while computationally cheaper, may cause the model to \textit{forget} the skills from earlier rounds of training. On the other hand, in addition to being more expensive, RT on merged datasets is possible only when the practitioner has access to the the training datasets from earlier rounds, which is not the case for many publicly available instruction-tuned LMs such as Mistral 7B \citep{jiang2023mistral} and Llama 3 \citep{llama3}. While we focus on publicly available mixes, this is an important consideration as new models and datasets are released.

As an alternative, we explore merging the parameters of models separately trained for individual skills---merging to learn---a family of approaches we refer to as ``parallel train then merge'' (PTM). This general idea is shared by several well-known methods (e.g., model patching, \citealp{Ilharco2022PatchingOM}; WiSE-FT, \citealp{wortsman2022robust}; task arithmetic, \citealp{ilharco2023editing}; TIES \citealp{yadav2023tiesmerging}; DARE, \citealp{yu2024language}; time vectors, \citealp{nylund2023time}). Unlike RT, PTM does not require access to the original training data, as one can separately train only on the new data, also making PTM a computationally cheaper option. Since PTM does not directly change the weights associated with previously learned tasks, it should allow the model to retain more of its original skills compared to other methods. PTM also enables the efficient addition of \textit{multiple} new skills to a single model. This work is the first to systematically explore PTM for instruction tuning.

We compare  the three methods for model patching (CFT, RT, and PTM) in terms of model performance as well as  computational cost. We experiment with adding three new skills: scientific literature understanding, coding, and refusing unsafe requests, to T\"ulu, a general-purpose instruction-tuned model, and evaluate the performance of the models resulting from each method on a suite of evaluation datasets representing general instruction-following skills as well as those that target the new skills being added. We find that:

\begin{itemize}
    \item When optimizing performance on the new skills, PTM achieves competitive performance with the best RT models, with a 50--95\% improvement in training efficiency.
    \item PTM preserves nearly all of the original model's general skills versus a 10--40\% drop for CFT, while achieving similar skill-specific performance.
    \item Setting the mixture weight proportional to the ratio of the number of training steps dedicated to the new skill being added is a good heuristic when held-out data is not available, resulting in good model performance on new skills while preserving general performance. We find that this heuristic can also be effective for adding \emph{multiple} new skills to a single model.
    
    \item For enabling safety-related refusals, a skill that can be at odds with general skills, PTM proves to be particularly effective compared to RT and CFT by improving unsafe refusal rates, preserving general skills, and reducing exaggerated refusals by 30--80\%.
\end{itemize}
Overall, we find that PTM is consistently a better choice than CFT for teaching instruction-tuned models new skills without compromising on general skills. We note that RT is not always a feasible option because the training datasets for many publicly available instruction-tuned models are not available, and even when it is, PTM generally offers comparable performance tradeoffs while being significantly cheaper.
\section{Problem Setup}
\label{sec:prob_setup}

We aim to add new, diverse behaviors to a general-purpose instruction-tuned model, while preserving the original model’s overall performance. We want to do so in a computationally efficient manner, without increasing inference cost. Defined concretely, we want to use a new skill-specific dataset $D$ to improve the performance of a general model $\theta_G$, trained on general data $G$, on an evaluation set for new skills $E_D$, without losing performance on a general set $E_G$, while minimizing the computational requirements.

We describe the methodology and training complexity of three methods for adding new skills to preexisting models: continued finetuning, retraining from scratch, and model merging. We measure training complexity in terms of how many training steps are required to create the pool of models we select from.

\subsection{Continued Finetuning on Target Skills} 
\label{sec:cft}

One straightforward method is to continue finetuning the instruction-tuned model on instruction data targeting the new skills, which is much cheaper than retraining from scratch. To determine the best amount of $D$ to train on, we try $n$ different subsamples $D_i$, each requiring $|D_i|$ training steps. In total, this requires $\sum_{i=1}^n |D_i|$ training steps.

We will see in Section~\ref{sec:overall_trends} that continued finetuning substantially degrades general skills.

\subsection{Retraining From Scratch}
\label{sec:rt}

Another method is to retrain the instruction-tuned model starting from the pretrained model, with the skill-specific data added into the original data mix. Since we care about retaining the general performance as well, the ratio of the amount of the skill-specific data to that of the original data needs to be determined carefully.  The ideal way to do this is to retrain with different data mixing ratios and then perform \textit{model selection} based on these models' performance on a held-out validation set.

While retraining should lead to competitive performance on both general and skill-specific evaluations, it is inefficient compared to other methods due to the need to retrain on the entire general mix for every training run, especially considering the model selection described above.

For a given subsample of the skill-specific data $D_i$, a single retraining run requires $|G| + |D_i|$ training steps. For $n$ data mix variations, the total is $n \cdot |G| + \sum_{i=1}^n |D_i|$. General instruction datasets tend to be larger than skill-specific datasets, containing hundreds of thousands to millions of instances \citep{ivison2023camels, OpenOrca, singh2024aya}, while many skill-specific datasets contain on the order of tens to a hundred thousand instances \citep{codealpaca, wadden2024sciriff, zheng2024opencodeinterpreter}, highlighting the overall expense of retraining.

Additionally, we note that retraining is \textit{not possible} in cases where the pretrained and instruction-tuned models have been released but the general instruction mix has \textit{not}, such as Llama 3 \citep{llama3}, Mistral 7B \citep{jiang2023mistral} and Gemma \citep{gemmateam2024gemma}. While we experiment with publicly available data, this is important for future work with the latest models and datasets.

\subsection{Parallel Training \& Merging}
\label{sec:model_merging}

We next describe the three model merging methods that we explore: \textbf{task arithmetic} \cite{ilharco2023editing}, \textbf{linear interpolation} \cite{rofin2022linear}, and \textbf{WiSE-FT} \cite{wortsman2022robust}. We consider these as instantiations of a general \textit{parallel-train-then-merge} (PTM) framework:\footnote{This name derives from ``branch-train-merge,'' a similar technique designed for pretraining \citep{li2022branchtrainmerge}.} training a base model on \textit{only} the skill-specific data $D$ to create the skill-specific model $\theta_D$, and then weight-space merging $\theta_D$ with a generalist model $\theta_G$ with weight $\omega$.

While retraining and continued finetuning require training multiple models to determine how much the skill-specific data should influence the general model, in PMT, this is accomplished through the mixture weighting parameter $\omega$. This means that the total training cost for PMT is $|D|$, dramatically lower than other methods.

The three PTM methods we consider correspond to different ways of training separately on the skill-specific data and incorporating it into the final model. While we describe all three methods below, we primarily focus on task arithmetic as we find that it is particularly adept at improving performance on new skills while preserving general skills. We compare all three directly in Section~\ref{sec:alternative_methods}.

\paragraph{Task Arithmetic}

For task arithmetic, we finetune the pretrained model $\theta_{\mathit{pre}}$ on all of the available skill-specific data $D$ to create $\theta_D$, and then subtract $\theta_{\mathit{pre}}$ to get the new task vector $\tau_D$:
\begin{equation}
    \tau_D = \theta_{D} - \theta_{\mathit{pre}}
\end{equation}
We then merge the task vector into the general-purpose instruction-tuned model $\theta_G$:
\begin{equation}
    \theta_{\mathit{final}} = \theta_G + \omega \cdot \tau_D,
\end{equation}
where $\omega$ is selected using held out data when available, or with a heuristic. Experimentally, we find that setting $\omega < 1.0$ is better than naively setting $\omega = 1$.

\paragraph{Linear Interpolation}

For linear interpolation, we create a general skill task vector $\tau_G$ by subtracting $\theta_{\mathit{pre}}$ from the instruction-tuned model:
\begin{equation}
    \tau_G = \theta_{G} - \theta_{\mathit{pre}}
\end{equation}
We then interpolate between the task vector $\tau_D$ and the general skill vector:
\begin{equation}
    \theta_{\mathit{final}} = \theta_{\mathit{pre}} + \omega \cdot \tau_D + (1-\omega) \cdot \tau_G
\end{equation}

\paragraph{WiSE-FT}

For WiSE-FT, we first continue to finetune $\theta_G$ on the new skill-specific dataset to create $\theta_{\mathit{CFT}}$. We then subtract  $\theta_G$ to create $\tau_{\mathit{CFT}}$:
\begin{equation}
    \tau_{CFT} = \theta_{\mathit{CFT}} - \theta_G
\end{equation}
We then add $\tau_{\mathit{CFT}}$ to the general model with weight $\omega$, downweighting the impact of CFT:
\begin{equation}
    \theta_{\mathit{final}} = \theta_G + \omega \cdot \tau_{\mathit{CFT}}
\end{equation}

Notably, WiSE-FT is especially suitable for models without publicly available pretrained checkpoints, as it does not require access to $\theta_{\mathit{pre}}$.

\section{Experimental Setup}

\subsection{Datasets \& Evaluations}

\label{sec:datasets-and-evals}

\begin{table*}
\begin{tabular}{lrrl}
\toprule
\textbf{Dataset}           & \textbf{\begin{tabular}[c]{@{}c@{}}Training\\Set Size\end{tabular}} & \textbf{\begin{tabular}[c]{@{}c@{}}Number\\of Evals\end{tabular}} & \textbf{Domain}                                \\
\midrule
T\"ulu V2 (modified) \citep{ivison2023camels}  & 275,464 & 5 & Collection of general instructions \\
\midrule
SciRIFF \citep{wadden2024sciriff}           & 61,349 & 9 & Scientific literature understanding \\
Safety (internal) & 66,161 & 4 & Prompt/refusal pairs \\
CodeFeedback \cite{zheng2024opencodeinterpreter} & 156,526 & 2 & Single-turn coding examples \\
\bottomrule
\end{tabular}
\caption{Summary of datasets used in this work.}
\label{tab:dataset_stats}
\end{table*}

We explore the trade-off between general and skill-specific performance across three sets of skills: \textbf{Science}, \textbf{Safety}, and \textbf{Coding}. We train general purpose models using a modified version of the T\"ulu V2 mix \citep{ivison2023camels}, and train skill-specific models with (1) SciRIFF \citep{wadden2024sciriff}, (2) a novel refusals dataset, and (3) CodeFeedback \citep{zheng2024opencodeinterpreter}. We additionally choose consistent sets of evaluations designed to capture either general or specialized skills. Datasets and relevant evaluations are described in greater depth below and in Table~\ref{tab:dataset_stats}.

\paragraph{General-purpose}

To train our general-purpose models, we use a modified version of the T\"ulu V2 mix. We removed the science subset (7.5k examples), CodeAlpaca (20k examples), and refusals identified by heuristics (23k examples), resulting in 275k total instances, to simulate a setting where the base model has plenty of room to improve in our target skills. We refer readers to \citealp{wang2023far} and \citealp{ivison2023camels} for more details on the original mix.

We evaluate general skills on a subset of the T\"ulu 2 evaluation suite to cover a broad range of skills: world knowledge (MMLU, \citealp{hendrycks2021measuring}), mathematics (GSM8K, \citealp{cobbe2021training}), open-ended generation (AlpacaEval, \citealp{alpaca_eval}), reasoning (Big Bench Hard, \citealp{suzgun2022challenging}), and truthfulness (TruthfulQA, \citealp{lin2022truthfulqa}). More details are available in \citealp{ivison2023camels}.

\paragraph{Science}

To train our skill-specific models on science, we use SciRIFF \citep{wadden2024sciriff}, an instruction dataset covering tasks like information extraction, question answering, and more for scientific literature understanding. The dataset covers scientific disciplines like biomedicine, artificial intelligence, and others. We refer readers to \citealp{wadden2024sciriff} for full details on the dataset.

We evaluate our models on science using SciRIFF's validation and test sets, measuring performance across nine held-out tasks not seen during training. Unless otherwise noted below, we report average \textit{validation} set scores during our analyses. We refer readers to \citealp{wadden2024sciriff} for more details on evaluations.

\paragraph{Safety}

To train our skill-specific models on safety, we use an internally developed safety dataset covering broad categories like harmful language, malicious uses, misinformation, and more. Each example is a potentially dangerous or harmful prompt paired with a refusal generated from GPT-4 \citep{openai2024gpt4}. A seed set of prompts were written by humans, and more prompts were generated based on this seed set by GPT-4. Refusals were collected by prompting GPT-4, and keeping responses that were classified as a refusal.

To evaluate our models on safety, we use four categories of safety evaluations in our experiments: toxicity (ToxiGen, \citealp{hartvigsen2022toxigen}), automated red teaming (HarmBench, \citealp{mazeika2024harmbench}), refusing unsafe prompts (XSTest Unsafe, \citealp{röttger2024xstest}), and exaggerated refusals (XSTest Safe). We normalize scores for these evaluations so 0.0 is the worst possible score and 100.0 is the best. For scores reported below, we report the average of the first three metrics and consider exaggerated refusals separately.

\paragraph{Coding}

To train our skill-specific models on coding, we use the single-turn subset of CodeFeedback \cite{zheng2024opencodeinterpreter} of instruction/code pairs.  We refer readers to \citealp{zheng2024opencodeinterpreter} for more details.

We evaluate our models on coding by taking the average of scores on HumanEval+ \citep{chen2021evaluating, liu2023code} and Mostly Basic Python Programs+ (MBPP+) \citep{austin2021program, liu2023code}. We sample with a temperature of 0.8 and report pass@10 metrics for both.

\subsection{Training Setup}

\paragraph{Settings}

We run all of our experiments on top of Llama 2 7B \citep{touvron2023llama}. We fully finetune all of our models for two epochs with a context length of 4,096 and a batch size of 128, and we follow the other hyperparameters used in \citealp{ivison2023camels}. Our general purpose model was created by training Llama 2 7B on our modified T\"ulu 2 mix, described in Section~\ref{sec:datasets-and-evals}. All models were trained on v3 TPUs using a fork of EasyLM\footnote{\url{https://github.com/hamishivi/EasyLM/tree/main}}. Full training details are available in Appendix~\ref{app:hyperparams}.

\paragraph{Methods For Adding New Tasks}

We compare three methods for add new skills to an instruction-tuned model, described in Section~\ref{sec:prob_setup}: \textbf{continued finetuning} (CFT) on skill-specific data, \textbf{retraining} (RT) from scratch on a mix of all of the general data and some amount of skill-specific data, and task arithmetic, a form of \textbf{parallel-train-then-merge} (PTM), by training the base pretrained model on the skill-specific data and adding the task vector directly to the instruction-tuned model.

We evaluate five checkpoints for each setting, varying the influence of the skill-specific data on the general model. For CFT and RT, we train on five different amounts of the skill-specific data. For PTM, we choose five different values for the mixture weight. We also compare PMT with linear interpolation and WiSE-FT in Section~\ref{sec:alternative_methods}.

Many instruction datasets do not have validation sets \cite{ivison2023camels, zheng2024opencodeinterpreter, OpenOrca, singh2024aya}, and thus how to select models is an open question. We take advantage of SciRIFF's validation set to select models in Section~\ref{sec:compute_cost-performance}. Otherwise, we select using heuristics.
\section{Results}
\subsection{Trade-off Between Computation Cost and Model Performance}
\label{sec:compute_cost-performance}
\begin{table}
\begin{tabular}{lrrr}
\toprule
\textbf{Model}                & \textbf{General} & \textbf{Science} & \textbf{\begin{tabular}[c]{@{}c@{}}Training\\ Steps\end{tabular}}           \\ \midrule
T\"ulu Only & 49.9 & 27.9 & - \\ \midrule
Best CFT & 33.7 & \textbf{40.6} & 1,005 \\
Best RT & \textbf{50.6} & 37.8 & 11,766 \\
\midrule
Best PTM  & 47.1 & 38.2 & \textbf{479} \\
\bottomrule
\end{tabular}
\caption{Performance on general and science test evaluations for the best continued finetuning (CFT), retraining from scratch (RT), and parallel-train-then-merge (PTM) models based on science validation performance. PTM shows equivalent science performance to the best RT model with slightly lagging general skills, while taking about 4\% as many training steps. PTM shows slightly lower science performance than CFT with an over 13 point gain on general skills, while taking less than half the total training steps. Detailed results are in Appendix~\ref{app:detailed_results}.}
\label{tab:science-coefficients}
\end{table}

In this section, we explore the performance and cost trade-offs when optimizing for performance based on held-out data. We look at improving scientific literature understanding through three methods: CFT, RT, and PTM.

For both CFT and RT, we experiment with the five different amounts of science data reported in SciRIFF \citep{wadden2024sciriff}, ranging from about 4k to about 61k training examples. For PTM, we train on all 61k training examples, and test five different values for the mixture weight $\omega$: 0.2, 0.4, 0.6, 0.8, and 1.0. We additionally report the total number of training steps required (with our batch size of 128) to create the models in each group to estimate the cost of each method. We then select the best model from each of these three methods based on SciRIFF validation set performance.

As shown in Table~\ref{tab:science-coefficients}, we find that PTM achieves strong performance on both general and science evaluations, with a fraction of the compute. PTM requires about \textbf{4\%} of the compute compared to RT, and slightly edges out the best RT model on science while being within a few points on general evaluations. PTM is also within a few points on science compared to the best CFT model, with a more than 13 point improvement in general skills and less than 50\% the compute.

\begin{table}
\begin{tabular}{lrrr}
\toprule
\textbf{Model} &  \textbf{\begin{tabular}[c]{@{}c@{}}\%$\Delta$\\ Gen. \end{tabular}}  & \textbf{\begin{tabular}[c]{@{}c@{}}\%$\Delta$\\ Spec. \end{tabular}} & \textbf{\begin{tabular}[r]{@{}r@{}}Training\\ Steps\end{tabular}}  \\
\midrule
Best CFT (Science) & --32.5 & 46.0 & 1,005 \\
Best RT (Science) & 1.37 & 39.1 & 11,766 \\
Best PTM (Science) & 1.30 & 26.3 & 479 \\
\midrule
Best CFT (Safety) & --40.1 & 98.9 & 1,551 \\
Best RT (Safety) & 0.66 & 89.6 & 12,311 \\
Best PTM (Safety) & --0.13 & 88.9 & 517 \\
\midrule
Best CFT (Coding) & --7.73 & 51.6 & 3,669 \\
Best RT (Coding) & 0.13 & 50.7 & 14,429 \\
Best PTM (Coding) & 1.43 & 33.3 & 1,223 \\
\midrule
Best CFT (Ex.~Ref.) & --85.1 & 98.9 & 1,551 \\
Best RT (Ex.~Ref.) & --39.9 & 87.2 & 12,311 \\
Best PTM (Ex.~Ref.) & --6.45 & 72.6 & 517 \\
\bottomrule
\end{tabular}
\caption{Absolute percentage change compared to the T\"ulu baseline on two evaluations: general and specialized for science, safety, and coding, and exaggerated refusals and safety for the exaggerated refusals rows. PTM  preserves general performance better than CFT and comparably to RT in all four settings, while requiring a fraction of the compute compared to either method. PTM also improves skill-specific performance in every scenario, and improves safety as much as RT while taking 4\% the compute. Detailed results in App.~\ref{app:detailed_results}.} 
\label{tab:model-selection}
\end{table}

\subsection{Model Performance Across Skills}

\paragraph{Overall Trends}
\label{sec:overall_trends}

We now look at overall trends when comparing PTM to CFT and RT. In Table~\ref{tab:model-selection}, we compare CFT, RT, and PTM across all three sets of skills, as well as for exaggerated refusals. We consider an equal number of models for each method to ensure a fair comparison: for CFT and RT, we perform five data mix trials for each dataset, and for PTM, we explore five evenly spaced values for $\omega$. We select models in each category based on their average percentage improvement over the baseline model in two dimensions: general skills and performance for science, safety, and coding, and exaggerated refusal compliance rate and safety for the exaggerated refusals rows.

\begin{figure*}[ht]
  \centering
  \includegraphics[width=1.0\textwidth]{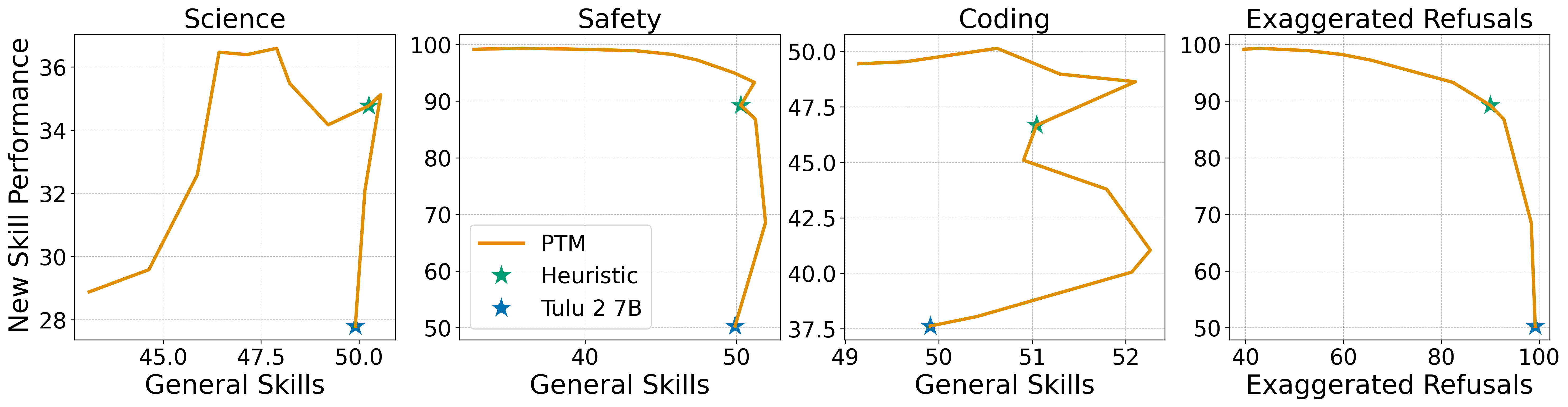}
  \caption{Trade-offs managed through $\omega$.  We highlight the point along each curve that corresponds to using our weighting heuristic, $\omega = \frac{|D|}{|G|}$. This point consistently achieves strong performance on all settings, without requiring held out data. We take advantage of PTM's negligible cost to test different mixture weights to plot 10 checkpoints from evenly spaced values of $\omega$ as well as the heuristic}
  \label{fig:ta-vs-heuristic}
\end{figure*}

\begin{table*}[t]
\centering
\begin{tabular}{lrrrrrr}
\toprule
\textbf{Model} & \textbf{General} & \textbf{Science} & \textbf{Coding} & \textbf{Safety} & \multicolumn{1}{l}{\begin{tabular}[r]{@{}r@{}}\textbf{Exaggerated} \\ \textbf{Refusals}\end{tabular}} & \multicolumn{1}{r}{\begin{tabular}[r]{@{}r@{}}\textbf{Additional}\\\textbf{Training} \\ \textbf{Steps}\end{tabular}}  \\ \midrule
T\"ulu Only & 49.9 & 27.8 & 37.6 & 50.3 & 99.2 & - \\ \midrule
CFT (All 3) & 40.3 & 37.9 & \textbf{58.2} & \textbf{99.8} & 16.0 & 2,219 \\
RT (All 3) & 50.1 & \textbf{39.2} & 57.9 & 95.0 & 37.2 & 4,732 \\
\midrule
PTM (All 3) & \textbf{51.1} & 26.6 & 45.3 & 84.0 & \textbf{93.2} & \textbf{0} \\
\bottomrule
\end{tabular}
\caption{PTM with all three skill-specific models improves general performance, while noticeably improving both coding and safety over the baseline. PTM also shows strong improvement on exaggerated refusals, with a \textbf{77 point gain} over CFT and a \textbf{56 point gain} over RT. However, science performance is adversely impacted, and CFT and RT achieve stronger coding and safety performance overall. If single-skill models have already been trained, three-skill PTM has \textbf{no} additional training cost, while both the CFT and RT models must be separately trained. Detailed results are in Appendix~\ref{app:detailed_results}.}
\label{tab:multidomains}
\end{table*}

From this table, we see a few trends. First, PTM preserves much more of the underlying model's general skills than CFT, which consistently suffers a substantial drop in general performance across all four settings. Second, for safety, PTM achieves comparable or better general \textit{and} safety-specific performance compared to RT. Finally, across all settings, PTM is substantially cheaper than both CFT and RT, requiring a fraction of the total training cost of either method.

\paragraph{PTM Mitigates Exaggerated Refusals}

As mentioned in Section~\ref{sec:datasets-and-evals}, when evaluating safety, we consider two metrics: our safety average, measuring how often a model correctly refuses to comply with a prompt, and exaggerated refusals, measuring how often the model complies with a prompt, written to be superficially similar to offensive or harmful prompts, that it \textit{should} comply with. In this section, we analyze the effect of using task vectors on exaggerated refusals.

We compare improvement over the baseline model for these metrics in Table~\ref{tab:model-selection}. By training a separate ``safety vector'' and applying it to an SFT model, we are able to achieve comparable, if not better, general and safety performance to CFT and RT, while \textbf{dramatically} improving compliance on exaggerated refusals. In other words, PTM makes the final model substantially better at safe but misleading prompts, improving performance on XSTest Safe by 30--50 points versus the other methods, while achieving similar safety and general performance.

\paragraph{Heuristics}

For settings without held-out data, such as our safety and coding datasets, we find that a consistently strong heuristic for selecting a model checkpoint is to 1) train the task vector on all of the available skill-specific data, and then 2) weight the vector with
\begin{equation}
    \omega = \frac{|D|}{|G|}.
\end{equation}
In Figure~\ref{fig:ta-vs-heuristic}, we plot the trade-off between general and skill-specific performance as we vary $\omega$ between $0.0$ and $1.0$. We highlight the point on each curve selected by our heuristic, and we see that it consistently selects a point on the curve that preserves most, if not all of the original model's general performance, while substantially improving skill-specific performance.

\subsection{Multiple New Skills}

We next investigate adding multiple new skills to a single model. In Table~\ref{tab:multidomains}, we report the results of merging all three skills into a single model using the heuristic, and compare with both CFT and RT using all of the available skill-specific data. We also report scores for using our heuristic for each skill individually.

We see that when we merge all three skills into a single model, we achieve best overall general and exaggerated refusals performance, once again \textbf{substantially} improving the latter over both CFT and RT. Additionally, we get very similar performance on both coding and safety when compared with single skill PTM models.
However, we also see a large drop in science performance. We take advantage of PTM's very low ablation cost to attempt to diagnose this issue. We investigate in Table~\ref{tab:2_domains} by merging each skill-specific model together pairwise, and find that the drop in science performance is most likely caused by interference between the coding and safety vectors, as demonstrated by the large difference in science performance between this merge, the baseline, and the two pairwise merges involving science. In Appendix~\ref{app:detailed_results}, we investigate if other model merging methods---designed to minimize interference---can help mitigate this issue.

\begin{table*}[t]
\centering
\begin{tabular}{lrrrrr}
\toprule
\textbf{Model} & \textbf{General} & \textbf{Science} & \textbf{Coding} & \textbf{Safety} & \multicolumn{1}{l}{\begin{tabular}[r]{@{}r@{}}\textbf{Exaggerated}\\ \textbf{Refusals}\end{tabular}} \\ \midrule
T\"ulu Only & 49.9 & 27.8 & 37.6 & 50.3 & 99.2 \\ \midrule
PTM (Science) & 50.3 & 34.8 & 36.1 & 50.9 & 98.0 \\
PTM (Safety) & 50.3 & 25.6 & 36.9 & 89.3 & 90.0 \\
PTM (Coding) & 51.0 & 24.9 & 46.7 & 50.5 & 98.0 \\
\midrule
PTM (Science and Safety) & 50.8 & 31.6 & 38.5 & 89.1 & 89.6 \\
PTM (Science and Coding) & 51.3 & 32.1 & 45.5 & 49.5 & 98.4 \\
PTM (Safety and Coding) & 52.1 & 18.8 & 45.2 & 85.0 & 92.4 \\
\bottomrule
\end{tabular}
\caption{We take advantage of the efficiency of PTM to attempt to diagnose the degraded science performance in the three-skill PTM results shown in Table \ref{tab:multidomains}, with no additional training cost. By merging each specialized model pairwise, we see that safety and coding together show a large drop in science performance compared to the single skill and baseline models, suggesting that the drop in science performance in the three skill model is caused by interference between the other two skills.}
\label{tab:2_domains}
\end{table*}

\subsection{Alternative PTM Methods}
\label{sec:alternative_methods}

The first alternatives we explore are the two other relative weighting schemes described in Section~\ref{sec:prob_setup}: linear interpolation and WiSE-FT. 

\begin{figure}[ht]
  \centering
  \includegraphics[width=20em]{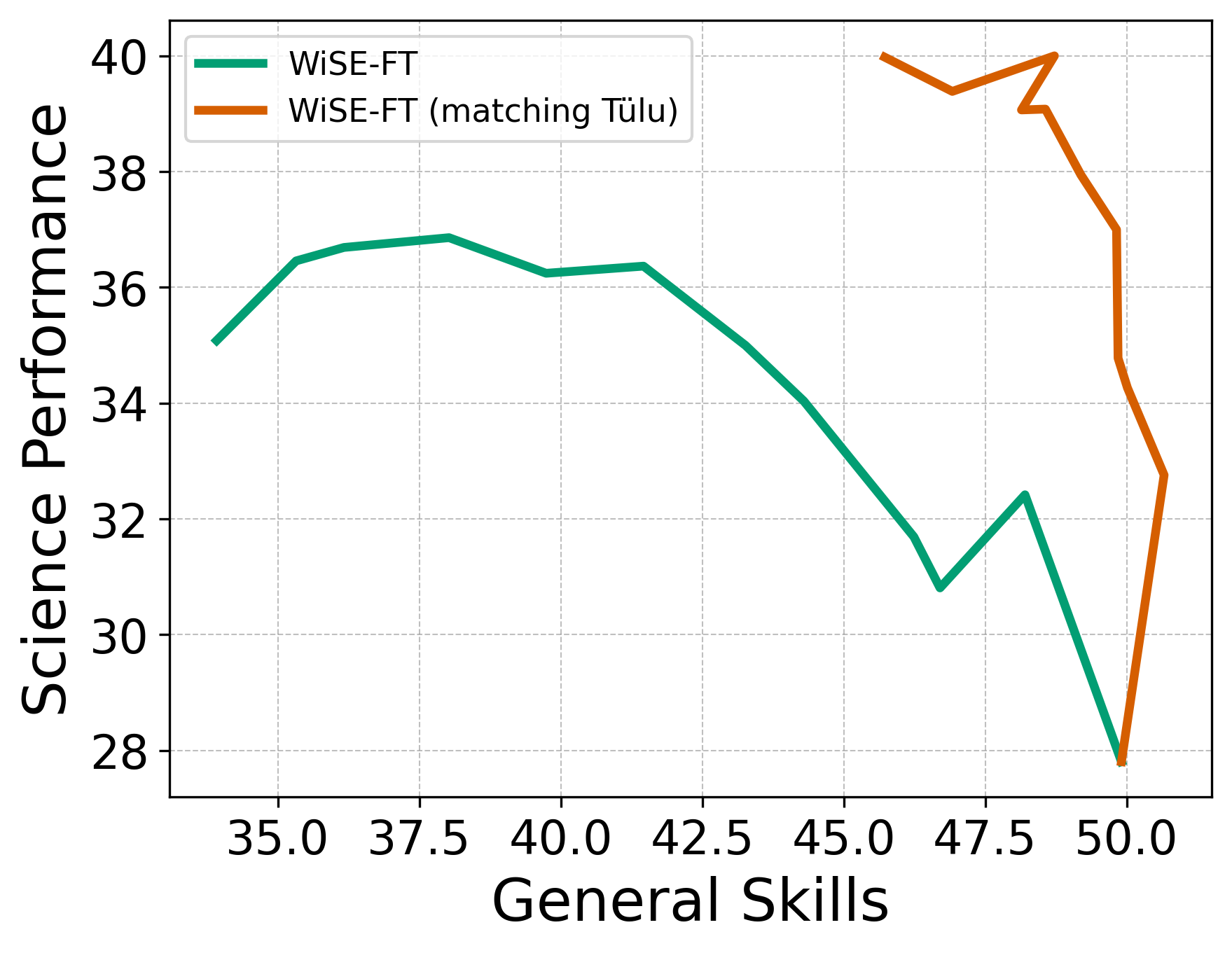}
  \caption{WiSE-FT performance on all of SciRIFF vs.~all of SciRIFF mixed with a matching amount of T\"ulu data. A matching amount of general data in the mix leads to an improvement in skill-specific performance and a much smaller degradation in general skills.}
  \label{fig:science_task_arithmetic_interference}
\end{figure}

\begin{figure*}[ht]
  \centering
  \includegraphics[width=1.0\textwidth]{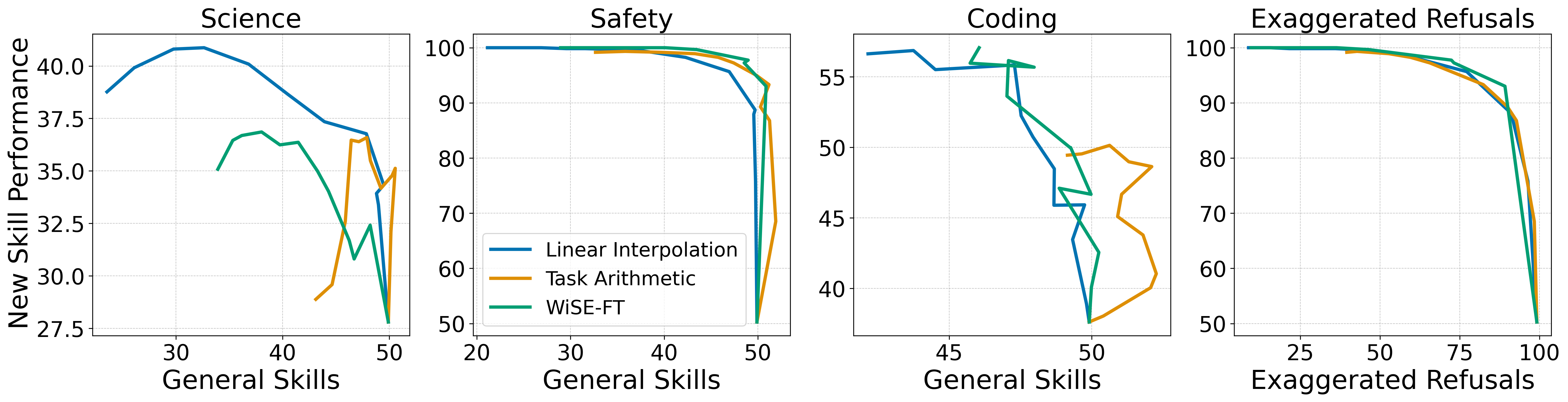}
  \caption{Plotting three PTM methods for each scenario. Both linear interpolation and WiSE-FT can achieve very strong domain-specific performance, at the cost of general performance and exaggerated refusals. While task arithmetic also improves in skill-specific performance, it preserves much more of the general skills.}
  \label{fig:3-weighting-strategies}
\end{figure*}

We compare all three methods for all three settings in Figure~\ref{fig:3-weighting-strategies}, using all of the skill-specific data in every scenario. We see that both linear interpolation and WiSE-FT can achieve strong skill-specific performance, and occasionally best overall, but consistently experience dramatic degradation in general performance compared to PTM.

For linear interpolation, as one vector is more highly weighted the other is downweighted, showing clear trade-offs between both sets of evaluations as we vary $\omega$.

For WiSE-FT, the findings seem strange at first glance. Why does continuing to finetune a strong generalist model on a specific skill degrade general performance \textit{more} than adding a task vector, which was trained in parallel on the base model? We hypothesize that this is caused by the differences between the skill-specific and general training data distributions.
In Figure~\ref{fig:science_task_arithmetic_interference}, we compare WiSE-FT trained on only science and WiSE-FT train on a mixture of science and a matching subsample of T\"ulu data. We see that this new mixture, which by construction is more similar to the general instruction data distribution already seen by the base model, achieves stronger skill-specific \textit{and} general performance compared to standard WiSE-FT. We leave it to future work to explore two follow up questions from this result:

\begin{enumerate}
    \item How much general data is needed during CFT to preserve general performance?
    \item When the base mix is not publicly available, is it possible to use data from a \textit{different} general distribution to preserve general performance?
\end{enumerate}

\section{Related Work}

\paragraph{Mitigating Forgetting During Finetuning} The problem of reduced generality in language models during finetuning, and more generally during continual learning is a fundamental problem in gradient based learning and has been widely studied in prior work~\citep{mccloskey1989catastrophic,goodfellow2013empirical,luo2023empirical}. Several techniques have been proposed to mitigate this issue, including regularization to minimize overfitting~\citep{ahn2019uncertainty,lee2019mixout} and recalling or replaying prior knowledge during training~\citep{kirkpatrick2017overcoming,röttger2024xstest,chen2020recall}. More recently, it has been shown that parameter-efficient learning methods, particularly low-rank adaptation~\citep{hu2021lora} forget less than conventional full-parameter finetuning methods. Unlike most of these methods, model merging does not require access to the original training data.

\paragraph{Model Editing}
% \pradeepd{Consider shortening these descriptions since you've introduced them earlier}
As discussed earlier,  \citet{Ilharco2022PatchingOM} introduced an approach that used model interpolation between a pretrained model and a model fine-tuned on a downstream task. This is conceptually similar to our work here (though we focus on language), and one of the foundational papers that spawned a variety of work.
\citet{ilharco2023editing} introduced ``task vectors'', which are calculated as the difference (in weight space) between a pretrained model and that same model after finetuning.
\citet{wortsman2022robust} introduced WiSE-FT, an approach of linearly interpolating between models, which they found to lead to increased distributional robustness.
\citet{wortsman2022model} introduced ``model soups'', made by weight-space averaging multiple models trained on the same dataset (using different hyperparameters, amounts of training data, etc.).
\citet{nylund2023time} introduced ``time vectors'', showing that \textit{temporal} information can also be created and applied to new models, similar to task vectors.

\section{Conclusions}

We explore the effectiveness of PTM for adding new skills to instruction-tuned models. We find that PTM is an efficient and effective method of augmenting preexisting models, enabling the addition of new skills with a fraction of the compute required compared to other common methods. In addition, we find that PTM achieves much better trade-offs between model safety and capability over other common methods tested. Finally, we report heuristics for selecting merging coefficients when held out data is not available, and find that these strategies together enable the addition of multiple skills into a single model with no additional training required.
\section*{Limitations}

While we aim to be comprehensive in our experiments, focusing on specific skills is inherently limiting. Instruction-tuning is also one step in a larger, constantly evolving adaptation framework, and this work does not test models that have undergone processes such as reinforcement learning from human feedback after instruction-tuning. Additionally, our evaluations, while covering a broad set of capabilities, do not capture the full set of abilities models can exhibit, such as general reasoning, or more abstract concepts such as helpfulness.

\section*{Acknowledgments}
Research supported with Cloud TPUs from Google’s TPU Research Cloud (TRC). PWK is supported by the Singapore National Research Foundation and the National AI Group in the Singapore Ministry of Digital Development and Innovation under the AI Visiting Professorship Programme (award number AIVP-2024-001). This research is also supported in part by Darpa ITM grant ONR N00014-24-1-2207

\bibliography{custom}

\appendix

\section{Training Details}

\label{app:hyperparams}

\subsection{Compute}

All of our models were trained on v3-128 TPUs on the Google TPU Research Cloud, and models were merged with the publicly available mergekit \cite{goddard2024arcee} toolkit.

\subsection{Hyperparameters}

We follow the hyperparameters used in \citealp{ivison2023camels}:

\begin{itemize}
    \item Precision: BFloat16
    \item Epochs: 2
    \item Weight decay: 0
    \item Warmup ratio: 0.03
    \item Learning rate: 2e-5
    \item Max. seq. length: 4,096
    \item Effective batch size: 128
\end{itemize}

\section{Other Results}
\label{app:detailed_results}

\paragraph{Other Merging Algorithms}

We also briefly experiment with two merging algorithms that aim to minimize interference between models: TIES \citep{yadav2023tiesmerging} and DARE \cite{yu2024language}. We compare these in Table~\ref{tab:multidomains-merging-methods} and find that they do not improve performance over weighted averaging.

\begin{table*}[t]
\centering
\begin{tabular}{lrrrrr}
\toprule
\textbf{Model} & \textbf{General} & \textbf{Science} & \textbf{Coding} & \textbf{Safety} & \multicolumn{1}{l}{\begin{tabular}[r]{@{}r@{}}\textbf{Exaggerated}\\ \textbf{Refusals}\end{tabular}} \\
\midrule
PTM (All 3) & 51.1 & 26.6 & 45.3 & 84.0 & 93.2 \\
TIES PTM (All 3) & 51.2 & 28.0 & 44.5 & 82.7 & 92.8 \\
DARE PTM (All 3) & 49.9 & 24.7 & 45.4 & 84.7 & 92.4 \\
\bottomrule
\end{tabular}
\caption{Results of using our heuristic to merge all three task vectors into a single model with standard weighted averaging PTM, TIES \citep{yadav2023tiesmerging}, and DARE \citep{yu2024language}. TIES and DARE also do not mitigate interference relative to the base method on science, and three methods show similar performance across all skills, showing that other popular merging algorithms do not mitigate interference in this setting.}
\label{tab:multidomains-merging-methods}
\end{table*}

% Finally, we merge all three datasets into a single model with TIES \citep{yadav2023tiesmerging} and DARE \cite{yu2024language}, described in Section~\ref{sec:alternative_methods}, and compare with standard weighted averaging in Table~\ref{tab:multidomains-merging-methods}. We see that neither method substantially improves science performance, and with our heuristic, performance is very similar across all three algorithms.

\paragraph{Exaggerated Refusals vs General Skills}

\begin{figure}[ht]
  \centering
  \includegraphics[width=20em]{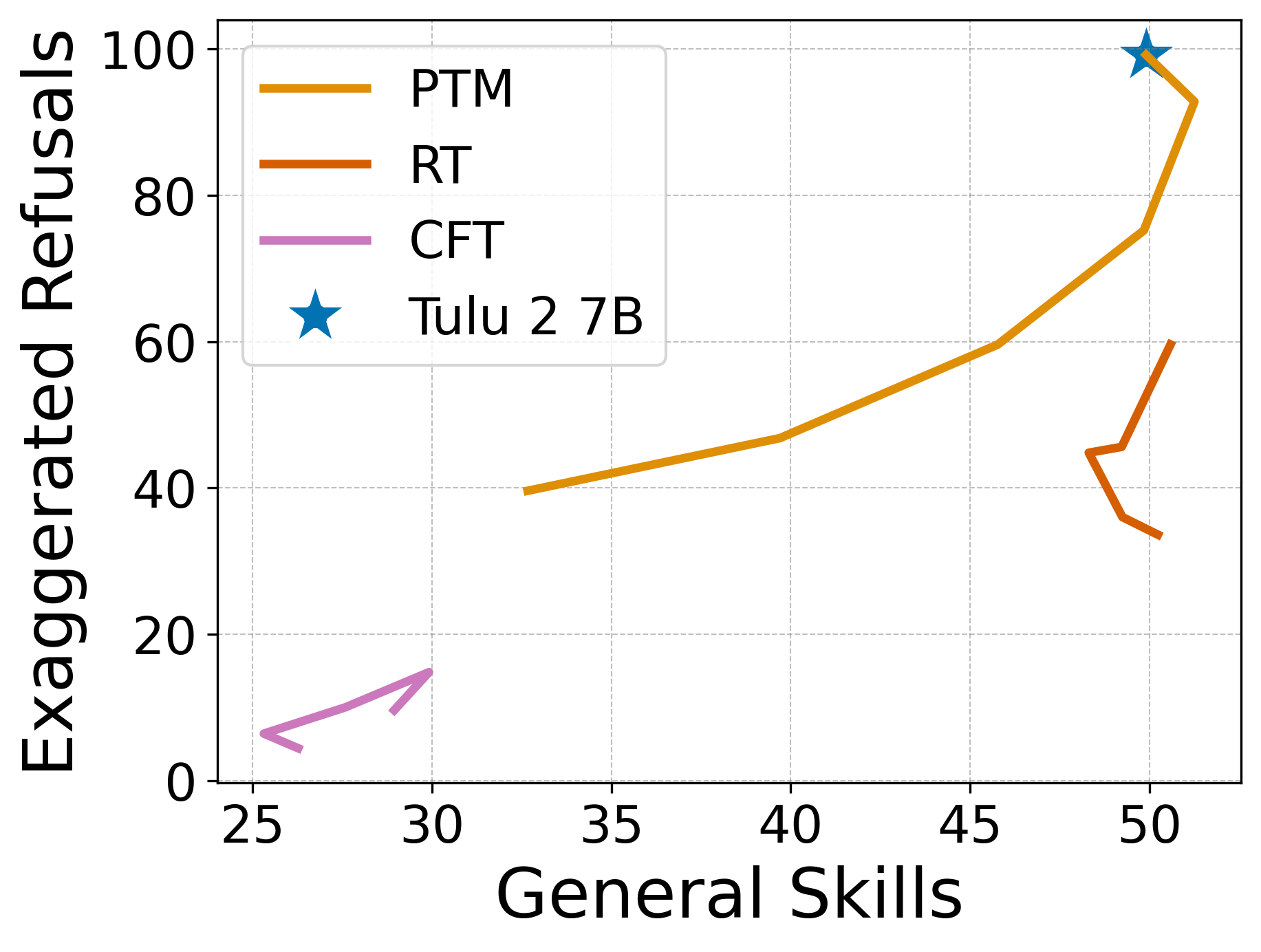}
  \caption{We show general skills versus exaggerated refusals, and show a clear relationship between the two skill sets. Additionally, for the same general performance, PTM achieves much higher exaggerated refusals compliance than RT and CFT.}
  \label{fig:tulu_vs_exaggerated_refusals}
\end{figure}

\begin{figure}[ht]
  \centering
  \includegraphics[width=20em]{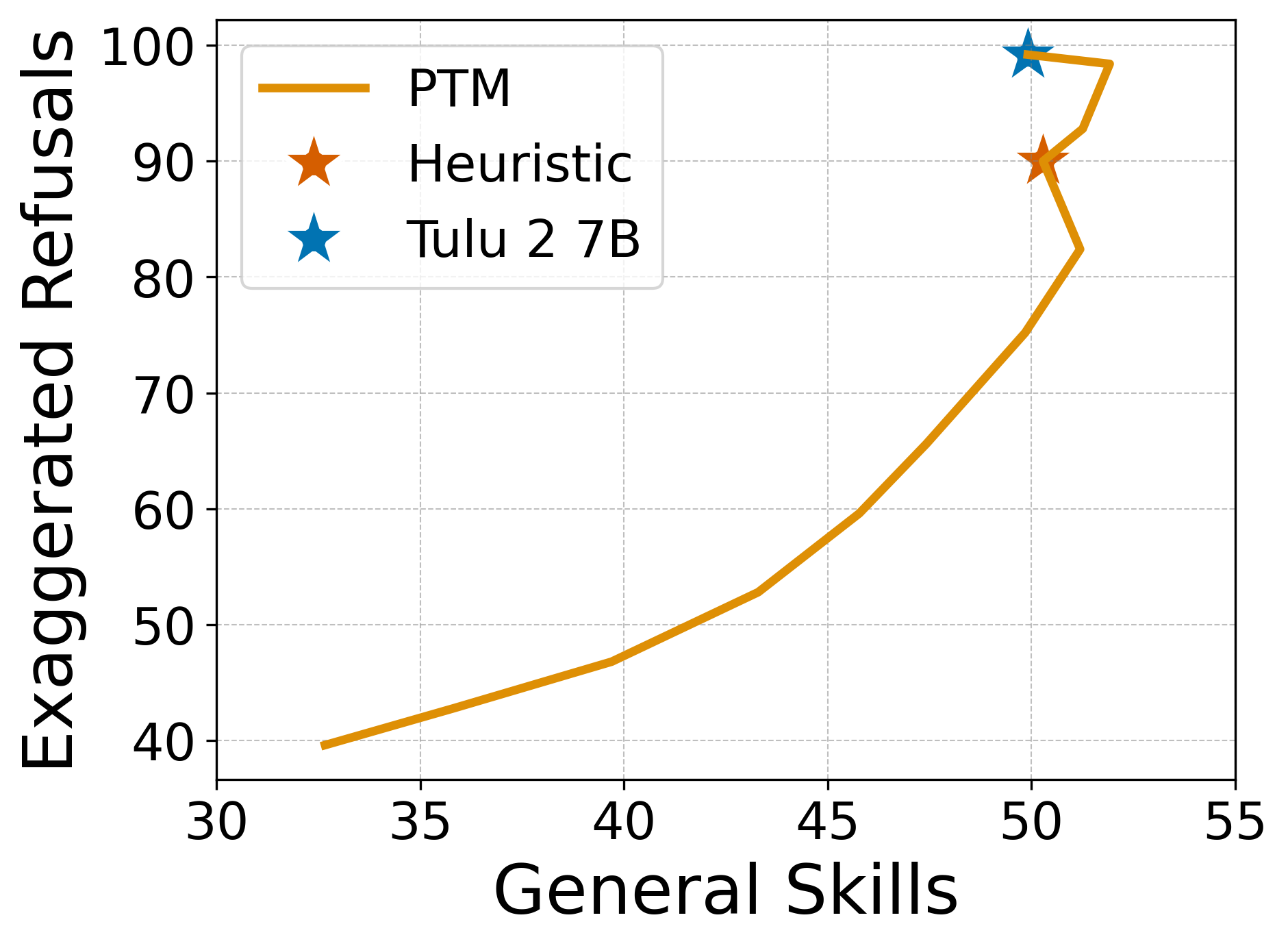}
  \caption{We show general skills versus exaggerated refusals, and highlight the point chosen by our heuristic, showing at most a small degradation in exaggerated refusal performance while preserving general skills.}
  \label{fig:tulu_vs_exaggerated_refusals-heuristic}
\end{figure}

We directly compare the trade-offs between general skills and exaggerated refusals in Figure~\ref{fig:tulu_vs_exaggerated_refusals} and Figure~\ref{fig:tulu_vs_exaggerated_refusals-heuristic}. We see that modifying the mixture weight $\omega$ shows a clear relationship between the two skills, and that PTM shows large improvements in exaggerated refusals over both CFT and RT

\begin{figure*}[ht]
  \centering
  \includegraphics[width=1.0\textwidth]{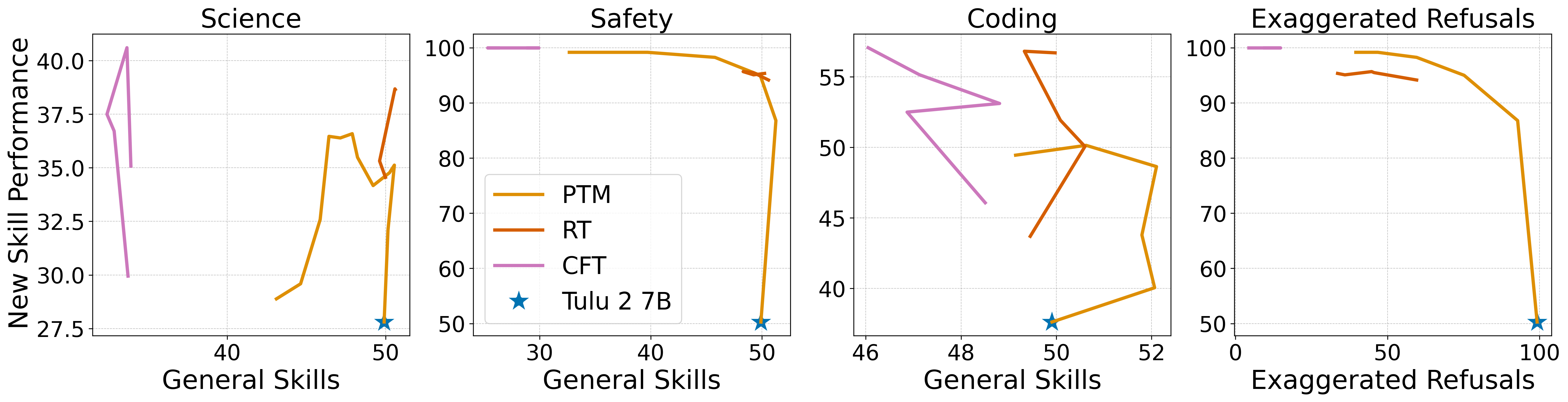}
  \caption{Curves showing general skills versus specialized skill performance for CFT, RT, and PTM across all three domains and exaggerated refusals. PTM consistently achieves strong specialized performance while preserving much more general performance compared to CFT. PTM also exhibits comparable performance to many RT checkpoints. In the case of exaggerated refusals, PTM shows a clear improvement over all other methods tested.}
  \label{fig:task_arithmetic_6_curves}
\end{figure*}
\section{Detailed Evaluations}
\label{app:detailed}

\begin{table*}
\begin{tabular}{lccccc}
\toprule
\textbf{Model} & \textbf{BBH} & \textbf{GSM8K} & \textbf{MMLU} & \textbf{TruthfulQA} & \textbf{Alpaca Eval} \\
\midrule
T\"ulu Only & 44.8 & 32.5 & 49.6 & 48.3 & 74.2 \\
\midrule
Best CFT & 45.1 & 22.0 & 46.0 & 38.3 & 17.0 \\
Best RT & 45.3 & 36.0 & 50.2 & 49.4 & 72.0 \\
\midrule
Best PTM & 44.5 & 26.5 & 48.5 & 49.6 & 66.7 \\
\bottomrule
\end{tabular}
\caption{Individual general evaluation results for Table~\ref{tab:science-coefficients}}
\label{tab:table-label1}
\end{table*}

\begin{table*}
{\small
\begin{tabular}{lccccccccc}
\toprule
\textbf{Model} & \textbf{BioASQ} & \textbf{BioRED} & \textbf{DiSCoMaT} & \textbf{Ev. Inf.} & \textbf{MultiCite} & \textbf{MUP} & \textbf{QASPER} & \textbf{SciERC} & \textbf{SciFact} \\
\midrule
T\"ulu Only & 44.1 & 33.5 & 22.8 & 15.6 & 35.2 & 19.8 & 15.1/23.7 & 14.4 & 54.5/38.5 \\
\midrule
Best CFT & 37.4 & 62.1 & 62.0 & 5.2 & 52.9 & 15.5 & 21.8/43.9 & 37.8 & 65.7/50.0 \\
Best RT & 34.9 & 50.5 & 58.6 & 10.2 & 48.7 & 16.8 & 15.3/45.6 & 29.3 & 65.4/47.3 \\
\midrule
Best PTM & 32.6 & 58.8 & 47.2 & 11.7 & 51.6 & 19.0 & 18.6/42.9 & 30.1 & 66.8/48.5 \\
\bottomrule
\end{tabular}
}
\caption{Individual science evaluation results for Table~\ref{tab:science-coefficients}}
\label{tab:table-label2}
\end{table*}

\begin{table*}
\begin{tabular}{lccccc}
\toprule
\textbf{Model} & \textbf{BBH} & \textbf{GSM8K} & \textbf{MMLU} & \textbf{TruthfulQA} & \textbf{Alpaca Eval} \\
 \midrule
T\"ulu Only & 44.8 & 32.5 & 49.6 & 48.3 & 74.3 \\
\midrule
Best CFT (Science) & 45.1 & 22.0 & 46.0 & 38.3 & 17.0 \\
Best RT (Science) & 45.3 & 36.0 & 50.2 & 49.4 & 72.0 \\
Best PTM (Science) & 47.9 & 33.5 & 49.2 & 49.7 & 72.5 \\
\midrule
Best CFT (Safety) & 34.7 & 24.5 & 46.7 & 39.0 & 4.7 \\
Best RT (Safety) & 45.4 & 34.5 & 49.8 & 48.0 & 73.5 \\
Best PTM (Safety) & 41.2 & 31.0 & 48.4 & 65.7 & 62.9 \\
\midrule
Best CFT (Coding) & 43.1 & 28.0 & 48.2 & 44.9 & 66.0 \\
Best RT (Coding) & 45.1 & 30.5 & 49.9 & 51.4 & 73.0 \\
Best PTM (Coding) & 41.9 & 24.5 & 47.8 & 58.4 & 80.6 \\
\midrule
Best CFT (Ex. Ref.) & 34.7 & 24.5 & 46.7 & 39.0 & 4.7 \\
Best RT (Ex. Ref.) & 46.9 & 32.5 & 49.8 & 49.3 & 74.4 \\
Best PTM (Ex. Ref.) & 48.4 & 29.5 & 48.5 & 58.9 & 71.0 \\
\bottomrule
\end{tabular}
\caption{Individual general evaluation results for Table~\ref{tab:model-selection}}
\label{tab:table-label3}
\end{table*}

\begin{table*}
{\small
\begin{tabular}{lccccccccc}
\toprule
\textbf{Model} & \textbf{BioASQ} & \textbf{BioRED} & \textbf{DiSCoMaT} & \textbf{Ev. Inf.} & \textbf{MultiCite} & \textbf{MUP} & \textbf{QASPER} & \textbf{SciERC} & \textbf{SciFact} \\ \midrule
T\"ulu Only & 44.5 & 33.2 & 22.8 & 15.9 & 35.3 & 19.7 & 15.1/23.7 & 13.1 & 54.3/28.2 \\ \midrule
Best CFT & 37.4 & 62.4 & 62.0 & 5.2 & 52.9 & 15.5 & 21.9/43.9 & 37.6 & 65.7/42.1 \\
Best RT & 34.9 & 50.5 & 59.2 & 10.8 & 48.8 & 17.2 & 21.0/42.7 & 33.6 & 66.1/40.8 \\ \midrule
Best PTM & 44.8 & 50.2 & 49.7 & 14.8 & 36.0 & 20.1 & 16.6/29.1 & 28.9 & 61.2/35.0 \\ \bottomrule
\end{tabular}
}
\caption{Individual science evaluation results for Table~\ref{tab:model-selection}}
\label{tab:table-label4}
\end{table*}

\begin{table}
\begin{tabular}{lcc}
\toprule
\textbf{Model} & \textbf{HumanEval+} & \textbf{MBPP+} \\
\midrule
T\"ulu Only & 29.8 & 45.5 \\
\midrule
Best CFT & 58.4 & 55.7 \\
Best RT & 56.5 & 56.9 \\
Best PTM & 44.9 & 55.4 \\
\bottomrule
\end{tabular}
\caption{Individual coding evaluation results for Table~\ref{tab:model-selection}}
\label{tab:table-label5}
\end{table}

\begin{table*}
\begin{tabular}{lcccc}
\toprule
\textbf{Model} & \begin{tabular}[c]{@{}c@{}}\textbf{XSTest Safe}\\ \textbf{(Exaggerated Refusals)}\end{tabular} & \textbf{Harmbench} & \textbf{XSTest} \textbf{Unsafe} & \textbf{ToxiGen} \\
\midrule
T\"ulu Only & 99.2 & 44.2 & 69.0 & 37.6 \\
\midrule
Best CFT & 85.2 & 100.0 & 100.0 & 100.0 \\
Best RT & 40.4 & 83.7 & 99.0 & 97 \\
Best PTM & 7.2 & 75.2 & 89.5 & 95.6 \\
\bottomrule
\end{tabular}
\caption{Individual safety evaluation results for Table~\ref{tab:model-selection}}
\label{tab:table-label6}
\end{table*}

\begin{table*}
\begin{tabular}{lccccc}
\toprule
\textbf{Model} & \textbf{BBH} & \textbf{GSM8K} & \textbf{MMLU} & \textbf{TruthfulQA} & \textbf{Alpaca Eval} \\ \midrule
T\"ulu Only & 44.8 & 32.5 & 49.6 & 48.3 & 74.2 \\ \midrule
CFT (All 3) & 45.2 & 25.0 & 47.3 & 47.3 & 37.0 \\
RT (All 3) & 45.4 & 32.5 & 49.6 & 51.7 & 71.2 \\ \midrule
Best PTM & 37.8 & 27.5 & 47.9 & 65.9 & 76.3 \\ \bottomrule
\end{tabular}
\caption{Individual general evaluation results for Table~\ref{tab:multidomains}}
\label{tab:table-label11}
\end{table*}

\begin{table*}
{\small
\begin{tabular}{lccccccccc}
\toprule
\textbf{Model} & \textbf{BioASQ} & \textbf{BioRED} & \textbf{DiSCoMaT} & \textbf{Ev. Inf.} & \textbf{MultiCite} & \textbf{MUP} & \textbf{QASPER} & \textbf{SciERC} & \textbf{SciFact} \\ \midrule
T\"ulu Only & 44.5 & 33.2 & 22.8 & 15.9 & 35.3 & 19.7 & 15.1/23.7 & 13.1 & 54.3/28.2 \\ \midrule
CFT (All 3) & 27.8 & 54.5 & 61.2 & 7.0 & 46.2 & 16.6 & 21.7/44.8 & 35.0 & 61.0/40.9 \\
RT (All 3) & 36.2 & 54.7 & 58.2 & 12.3 & 49.1 & 17.4 & 19.1/43.9 & 33.3 & 61.5/45.8 \\ \midrule
PTM (All 3) & 39.1 & 49.3 & 14.4 & 12.4 & 34.3 & 20.9 & 8.1/9.2 & 24.2 & 54.0/26.9 \\ \bottomrule
\end{tabular}
\caption{Individual science evaluation results for Table~\ref{tab:multidomains}}
\label{tab:table-label10}
}
\end{table*}

\begin{table}
\begin{tabular}{lcc}
\toprule
\textbf{Model} & \textbf{HumanEval+} & \textbf{MBPP+} \\ \midrule
T\"ulu Only & 29.8 & 45.5 \\ \midrule
CFT (All 3) & 57.1 & 59.4 \\
RT (All 3) & 56.3 & 59.5 \\
PTM (All 3) & 39.1 & 51.4 \\ \bottomrule
\end{tabular}
\caption{Individual coding evaluation results for Table~\ref{tab:multidomains}}
\label{tab:table-label8}
\end{table}

\begin{table*}
\begin{tabular}{lcccc}
\toprule
\textbf{Model} & \textbf{\begin{tabular}[c]{@{}c@{}}XSTest Safe\\ Exaggerated Refusals\end{tabular}} & \textbf{Harmbench} & \textbf{XSTest Unsafe} & \textbf{ToxiGen} \\ \midrule
T\"ulu Only & 99.2 & 44.2 & 69.0 & 37.6 \\ \midrule
CFT (All 3) & 16.0 & 99.5 & 100.0 & 100.0 \\
RT (All 3) & 37.2 & 85 & 100.0 & 99.9 \\
PTM (All 3) & 93.2 & 72 & 86.0 & 94.0 \\ \bottomrule
\end{tabular}
\caption{Individual safety evaluation results for Table~\ref{tab:multidomains}}
\label{tab:table-label9}
\end{table*}

\end{document}